\documentclass[letterpaper]{article} % DO NOT CHANGE THIS
\usepackage{aaai25}  % DO NOT CHANGE THIS
\usepackage{times}  % DO NOT CHANGE THIS
\usepackage{helvet}  % DO NOT CHANGE THIS
\usepackage{courier}  % DO NOT CHANGE THIS
\usepackage[hyphens]{url}  % DO NOT CHANGE THIS
\usepackage{graphicx} % DO NOT CHANGE THIS
\usepackage{natbib}  % DO NOT CHANGE THIS AND DO NOT ADD ANY OPTIONS TO IT
\usepackage{caption} % DO NOT CHANGE THIS AND DO NOT ADD ANY OPTIONS TO IT
\usepackage{algorithm}
\usepackage{newfloat}
\usepackage{listings}
\DeclareCaptionStyle{ruled}{labelfont=normalfont,labelsep=colon,strut=off} % DO NOT CHANGE THIS
\floatstyle{ruled}
\newfloat{listing}{tb}{lst}{}
\floatname{listing}{Listing}
\usepackage{enumitem}
\usepackage{amsfonts}
\usepackage{amsmath}
\usepackage{multirow}

\usepackage{algpseudocode}
\usepackage{booktabs}
\usepackage{bm}
\usepackage{color}
\usepackage[rgb,dvipsnames]{xcolor}

\newcommand{\etal}{\emph{et al.}}

\def\x{\bm x}

\def\btheta{{\boldsymbol \theta}}
\def\bphi{{\boldsymbol \phi}}

\def\tphi{\tilde{\boldsymbol{\phi}}}
\def\hphi{\hat{\boldsymbol{\phi}}}

\newcommand{\gain}[1] {\scriptsize\color{OliveGreen}{{#1}}}

\title{CALLIC: Content Adaptive Learning for Lossless Image Compression}
\author {
    Daxin Li\textsuperscript{\rm 1}\footnote{Equal contribution. $\dagger$Corresponding author.},
    Yuanchao Bai\textsuperscript{\rm 1}$^{*}$,
    Kai Wang\textsuperscript{\rm 1},
    Junjun Jiang\textsuperscript{\rm 1},
    Xianming Liu\textsuperscript{\rm 1}$^\dagger$,
    Wen Gao\textsuperscript{\rm 2}
}
\affiliations {
    \textsuperscript{\rm 1}Faculty of Computing, Harbin Institute of Technology, Harbin\\
    \textsuperscript{\rm 2}Department of Computer Science and Technology, Peking University, Beijing\\
    hahalidaxin@stu.hit.edu.cn, yuanchao.bai@hit.edu.cn, cswk@stu.hit.edu.cn, jiangjunjun@hit.edu.cn, \\ csxm@hit.edu.cn, wgao@pku.edu.cn
}
\usepackage{bibentry}
\begin{document}

\maketitle

\begin{abstract}
Learned lossless image compression has achieved significant advancements in recent years. However, existing methods often rely on training amortized generative models on massive datasets, resulting in sub-optimal probability distribution estimation for specific testing images during encoding process.  To address this challenge, we explore the connection between the Minimum Description Length (MDL) principle and Parameter-Efficient Transfer Learning (PETL), leading to the development of a novel content-adaptive approach for learned lossless image compression, dubbed CALLIC. Specifically, we first propose a content-aware autoregressive self-attention mechanism by leveraging convolutional gating operations, termed Masked Gated ConvFormer (MGCF), and pretrain MGCF on training dataset. Cache then Crop Inference (CCI) is proposed to accelerate the coding process. During encoding, we decompose pre-trained layers, including depth-wise convolutions, using low-rank matrices and then adapt the incremental weights on testing image by Rate-guided Progressive Fine-Tuning (RPFT). RPFT fine-tunes with gradually  increasing patches that are sorted in descending order by estimated entropy, optimizing learning process and reducing adaptation time. Extensive experiments across diverse datasets demonstrate that CALLIC sets a new state-of-the-art (SOTA) for learned lossless image compression.
\end{abstract}


\section{Introduction}

In the realm of digital imaging, lossless image compression stands as a pivotal technology, playing a crucial role in efficiently managing and transmitting vast amounts of image data without sacrificing quality. This form of compression is essential in fields where precision and detail are paramount, such as medical imaging, remote sensing, and digital archiving. 
% The importance of lossless image compression cannot be overstated, as it ensures the integrity and authenticity of the original image, a critical factor in applications where the slightest alteration could lead to misinformation or misinterpretation. 
Furthermore, the pursuit of higher compression ratios in lossless encoding is of paramount importance. Achieving greater compression rates without loss of information not only conserves valuable storage space and reduces transmission time but also addresses the growing demand for handling larger and more complex images, such as those in high-resolution photography and 3D medical scans. Thus, advancements in lossless image compression techniques are instrumental in driving forward the capabilities and efficiency of various data-intensive sectors.

Traditional image compression methods like PNG, JPEG-LS~\cite{weinberger2000loco}, and JPEG-XL~\cite{alakuijala2019jpeg} use handcrafted algorithms to exploit the statistical properties of images. However, these methods often struggle to capture the complex and diverse distributions in raw images, limiting their compression performance. In recent years, deep learning-based techniques have revolutionized neural compression~\cite{balle2017end, cheng2020learned, he2022elic, bai2022towards, li2024groupedmixer, li2024semantic}, and lossless compression methods~\cite{mentzer2019practical,kingma2019bit, townsend2019hilloc, hoogeboom2019integer, ho2019compression, zhang2021ivpf, zhang2021iflow, ryder2022split,rhee2022lc, wang2023learning, bai2024deep, Zhang2024ArIBBPS, wang2024learning} have achieved state-of-the-art (SOTA) results by learning the intricate distribution of raw images. These approaches use likelihood-based generative models, such as autoregressive models, flow models, and variational autoencoders (VAEs), to predict probabilities for entropy coding, converting data into compact bitstreams. Despite their success, these models often rely on training with large datasets, which can lead to sub-optimal probability estimates for specific testing images. Each image's unique details, such as textures, lighting, or structures, pose challenges that need to be addressed to improve compression performance.

To address the aforementioned challenge, we propose CALLIC, a content-adaptive learned lossless image compression method that leverages the connection between the Minimum Description Length (MDL) principle and Parameter-Efficient Transfer Learning (PETL). Specifically, we introduce an effective and lightweight pre-trained autoregressive model with a masked gating mechanism to simulate content-aware self-attention, termed the Masked Gated ConvFormer (MGCF). The pre-trained MGCF identifies patterns common across various images, known in advance to both the encoder and decoder. To make MGCF a practical codec, we propose Cache then Crop Inference (CCI), which caches intermediate activations and convolves only on the cropped patches surrounding the activated positions during coding.
During encoding, we enhance the pre-trained MGCF by incorporating additional parameter-efficient weights, fine-tuning it to capture the unique characteristics of each individual image and encoding this as a model prompt into a bitstream. We begin by decomposing the linear and depth-wise convolution layers in MGCF using low-rank matrices. Then, we introduce Rate-guided Progressive Fine-Tuning (RPFT), which fine-tunes low-rank weights with progressively increasing patches that are sorted in descending order by estimated entropy for more efficient adaptation. The MDL principle is employed to jointly optimize the bitrates for incremental weights and the compressed image. Finally, the adapted MGCF is used to encode the image, which is then sent to the decoder along with the compressed incremental weights.

Our contributions are encapsulated as follows:
\begin{itemize}
\item By bridging the MDL principle with PETL, we propose CALLIC, a novel content-adaptive lossless image compression method, effectively addressing the amortization gap between training data and testing images.
\item We introduce MGCF, a content-aware autoregressive self-attention mechanism with convolutional gating operations. CCI is proposed to speed up the coding process.
\item We decompose the pre-trained weights, including depth-wise convolutions, using low-rank matrices and adapt the model to testing image with RPFT, efficiently fine-tuning weights with progressively increasing patches sorted in descending order by estimated entropy.
% to optimize learning process and reduce adaptation time.
\item Extensive experimental results demonstrate that CALLIC achieves a new SOTA performance in lossless image compression, significantly outperforming existing learned methods across various datasets.
\end{itemize}

\section{Related Work}

\subsection{Learned Lossless Image Compression}
Lossless image compression converts images into the fewest possible bits while maintaining lossless reconstruction. 
Based on different types of likelihood models, these methods are categorized into autoregressive models, flow models, and VAE models.

\textit{Autoregressive models} decompose data probability into conditional distributions via the chain of probabilities, using learnable models for these distributions. Oord \etal~\cite{van2016pixel} introduced PixelCNN, estimating the conditional distribution with masked convolution. Bai \etal~\cite{bai2021learning} developed an end-to-end framework using lossy compression for prediction and a serial autoregressive model for residual image compression. They later advanced this with a parallel model in \cite{bai2024deep} for faster coding. 
% Rhee \etal~\cite{rhee2022lc} built upon autoregressive model, introducing conditional dependencies among different illumination and color components, and between low and high frequency areas.

For \textit{flow models}, Ho \etal~\cite{ho2019compression} presented a local bit-back coding scheme to modify a continuous normalizing flow model for lossless compression. In contrast, discrete normalizing flows handle discrete data and employ invertible transformations for integer latent space, which is encoded with a simple prior distribution. Hoogeboom \etal~\cite{townsend2019hilloc} developed an integer flow (IDF) model which learned invertible transformations for compression. 
% Berg \etal~\cite{van2020idf++} further refined IDF including improving the network architecture. 
Zhang \etal~\cite{zhang2021ivpf} introduced an invertible volume-preserving flow (iVPF) model, and later, the iFlow model in \cite{zhang2021iflow}, featuring modular scale transforms and uniform base conversion systems.

\textit{VAE models} are divided into deterministic and stochastic posterior sampling methods. Deterministic methods, like L3C~\cite{mentzer2019practical}, impose constraints on the latent codes' posterior distribution. Stochastic methods, exemplified by BB-ANS~\cite{townsend2019practical}, employ stochastic latent code sampling, using the evidence lower bound (ELBO) for training. Kingma \etal~\cite{kingma2019bit} later proposed the Bit-Swap scheme based on hierarchical VAE models. Townsend \etal~\cite{townsend2019hilloc} introduced HiLLoC, another hierarchical latent lossless compression method. 
% Ryder \etal \cite{ryder2022split} proposed a split hierarchical VAE model, merging hierarchical VAE with autoregressive models to factorize the prior distribution and address the initial bits issue.

\subsection{Parameter-Efficient Transfer Learning}

Presently, PETL strategies can be categorized into three predominant groups.

\textit{Adapter Modules}, introduced by Houlsby \etal~\cite{houlsby2019parameter}, are small, trainable networks added to existing models. These modules adjust the model's internal representations to new tasks, keeping most original parameters unchanged.

\textit{Prompt Tuning} uses task-specific prompts added to input data. This approach guides the pre-trained model to produce task-relevant outputs without changing its parameters~\cite{li2021prefix,lester2021power}.

\textit{Low-Rank Adaptation}~\cite{hu2021lora} modifies a pre-trained network's weights with low-rank matrices, enabling task adaptation with few additional parameters.

There have been some works that combine learned lossy image compression with PETL, while they insert adapters into transform network~\cite{tsubota2023universal,shen2023dec}, mainly to solve performance degradation for out-of-domain images in transform coding framework. 

\section{Minimum Description Length \label{sec:mdl}}

The MDL principle suggests that the most effective model for data description is the one that maximizes the data compression efficiency~\cite{grunwald2005minimum, barron1998minimum}. A notable variant of this principle is the crude MDL or two-stage MDL principle, which involves a two-pass coding process. First, the process entails selecting and encoding the optimal model into a bitstream. Second, data is compressed using the chosen model. Given random variable $\x$'s drawn from a probability distribution $p(\x)$, the probability density function of $\x=(x_1, x_2, \cdots, x_n)$ is defined as $p(x)=p(x_1, x_2, \cdots, x_n)$. The model (a probability density function) $q$ is selected from a countable collection $\mathcal{Q}$ by minimizing the total codelength: 
\begin{equation}
    \mathcal{L}_{\text{2-stage}}(\x) = \min_{q\in \mathcal{Q}} L(q)+L_q(\x).
\end{equation}
In this framework, $L(q)=\log \frac{1}{w(q)}$ denotes the codelength for the model (reflecting model complexity) and $w(q)$ represents a prior probability mass function on $q\in Q$. $L_q(\x)=\log \frac{1}{q(\x)}$ quantifies the codelength for data $x$ under model $q$. 

In learned lossless image compression, the image $\x$ is decomposed into a sequence of variables, enabling its representation as a set of non-independent data points $x_1, x_2, \cdots, x_n$. Each variable's unknown real distribution is estimated with a generative model, and entropy coding tools are used to encode each variable based on the estimated distribution. 
The generative model is trained by minimizing the negative log-likelihood on a training dataset. During encoding, both the encoder and decoder are aware of the trained model (ignoring $L(q)$), and only the bitstream of image (of length $L_q(\x)$) needs to be transmitted.
Nevertheless, the generative model with learned amortized parameters on the whole training dataset usually results in sub-optimal probability model $q$ for specific testing images.
Each testing image manifests its own unique details, such as textures, lighting or structures, stemming from diverse shooting targets and imaging devices. 
This leaves room for improvement by closing the gap between training and testing images.

To address this challenge, this paper incorporates the two-stage MDL principle into learned lossless image compression. 
Initially, we propose to fine-tune the pre-trained generative model to a single image and encode the fine-tuned model into a bitstream. 
However, fine-tuning and encoding the entire generative model is extremely inefficient in terms of both computation and storage, necessitating a more efficient and scalable approach for the image-specific adaptation.
Inspired by recent PETL, we fix the pre-trained generative model but add a small, trainable incremental weights to adjust the model's internal representations to each testing image.
Instead of encoding the entire fine-tuned model, the pre-trained generative model is known by both the encoder and decoder, and only the incremental weights need to be encoded to a bitstream.
We then compress the testing image using the adapted generative model, and transmit the compressed image together with the incremental weights to the decoder.
In order to maximize the coding efficiency, we leverage the MDL principle to jointly optimize the bitrates for both the incremental weights and the compressed image:
\begin{equation}
    \min_{\bm{\phi}} L(\bm{\phi})+\log \frac{1}{q(\x;\bm{\theta}, \bm{\phi})},
    \label{eq:loss_function}
\end{equation}
where $L(\bm{\phi}) = \log \frac{1}{w(\bm{\phi})}$ indicates the codelength required to encode the incremental weights $\bm{\phi}$, assumed to follow a prior distribution $w(\bm{\phi})$. 
The $\log \frac{1}{q(\x;\bm{\theta},\bm{\phi})}$ quantifies the codelength for current image $\x$ under the adapted model $q(\x;\bm{\theta}, \bm{\phi})$ incorporating both the pre-trained parameter $\bm{\theta}$ and the incremental weights $\bm{\phi}$.
The expected length of the two-stage code in relation to the actual data distribution $p$ is:
\begin{equation}
\label{eq:d2_m_kl}
    L(\bm{\phi}) + \mathbb{E}_{p}\left[\log \frac{1}{q(\x; \btheta, \bm{\phi})}\right] = L(\bm{\phi}) + D_{KL}(p\| q) + H(p),
\end{equation}
where $D_{KL}(p\| q)$ represents the Kullback-Leibler divergence between $p$ and $q$, and $H(p)$ is the entropy of $p$. 

A bound $R_n(p)$ on the expected redundancy between the optimization target and the real entropy is derived as follows:
\begin{equation}
    \begin{aligned}
% & E_{p}\left[\mathcal{L}_{\text { 2-stage }} \left(\x\right)-\log 1 / p\left(\x\right)\right] \\
& E_{p}\left[\min _{\bphi}\left\{L(\bphi)+ D_{KL}(p\| q)\right\}\right] \\
& \leq \min _{\bphi} E_{p}\left[L(\bphi)+ D_{KL}(p\| q) \right] \\
& =\min_{\bphi}\left\{L(\bphi)+D_{KL}(p \| q)\right\}=R_{n}(p) .
\end{aligned}
\end{equation}
The inequality holds using Jensen’s inequality since minimum is a concave function.
The bound $R_n(p)$ reflects the trade-off between estimation precision and model complexity in two-stage coding, as a more precise model requires more bits $L(\bphi)$ to describe the model, but exhibits a smaller divergence $D_{KL}(p\|q)$ with the true probability distribution.
%In encoding, a two-stage coding process is followed. The first stage involves optimizing local parameters for the current image and encoding them using a prior weight distribution $w(\bphi)$. The second stage entails encoding the image with the full model. 

This theoretical analysis paves the way for the following design of our content-adaptive learned lossless image compression, CALLIC. 
We approach content-adaptive lossless image compression from two aspects: 
\textbf{ 1) designing an effective and lightweight pre-trained generative model for content-aware structures
% , which achieves improved amortized performance by learning from a training dataset; 
2) developing an efficient fine-tuning method to adjust the pre-trained model to each testing image, optimizing the MDL in \eqref{eq:loss_function}. }
The subsequent sections will explore these two aspects.

\begin{figure*}[t]
 \centering
 \includegraphics[scale=0.29]{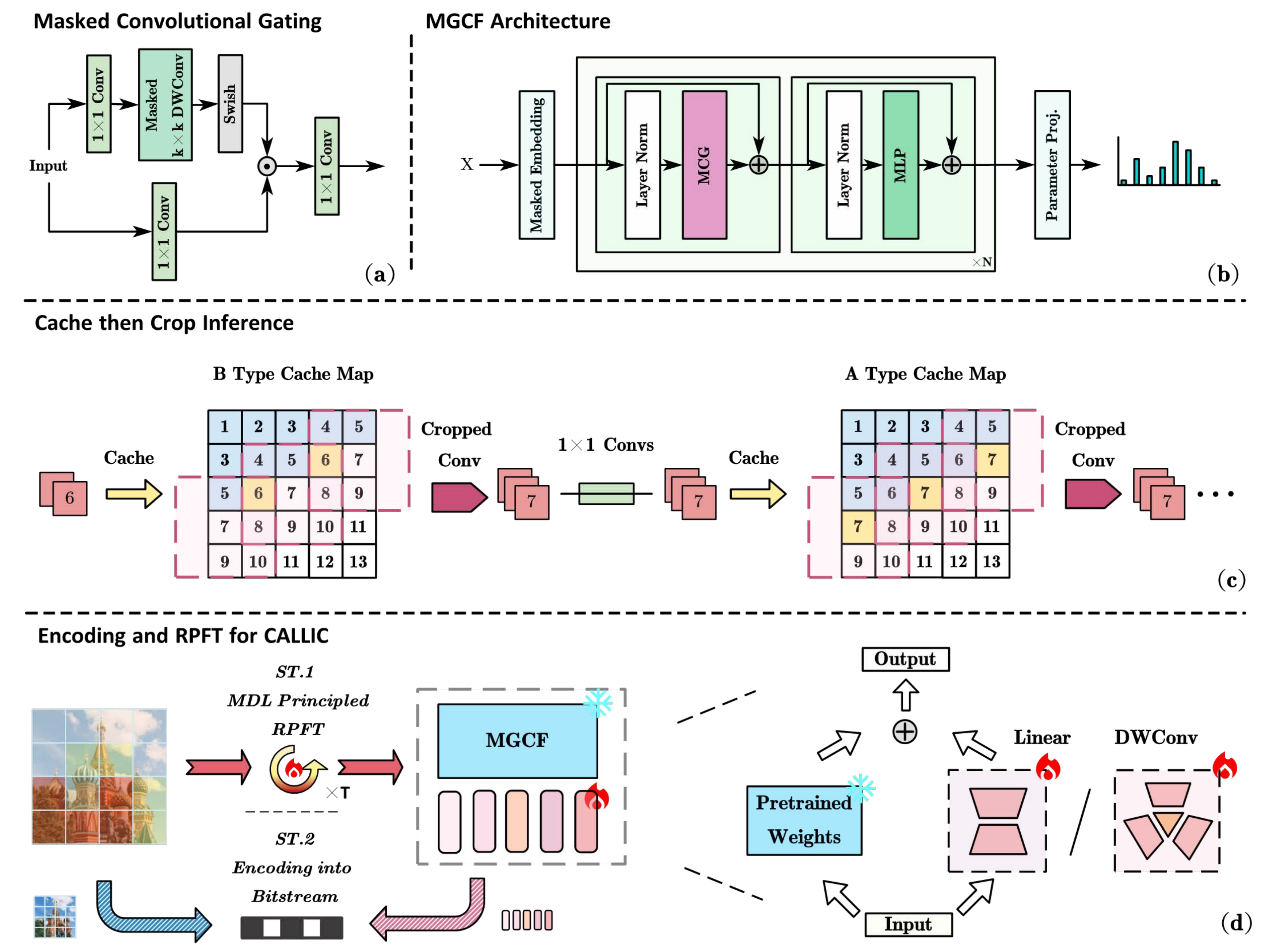}
 % \caption{Overview of key components in CALLIC. }
 \caption{Overview of the proposed architectures and mechanisms in CALLIC: We first simplify the attention mechanism using a content-adaptive convolutional gating mechanism (a). Building on this, we introduce the Masked Gated ConvFormer (MGCF) (b). To accelerate the coding process, we propose Cache then Crop Inference (CCI) (c), which caches activations before masked convolution layers and then performs convolution on cropped features at coding positions. For instance-level content adaptivity, we fine-tune the pre-trained model on the test image. This involves adding learnable parameters through low-rank decomposition of MGCF layers (Right), fine-tuning these additional parameters with the proposed Rate-guided Progressive Fine-tuning (RPFT), and encoding them into the bitstream along with the image (Left). }
\label{fig:network}
% \vspace{-8pt}
\end{figure*}

\section{CALLIC Method}
\subsection{Pre-trained Content Adaptive Model}
\noindent
\subsubsection{Masked Convolutional Gating Mechanism}
Transformers have empowered many fields~\cite{vaswani2017attention, dosovitskiy2020image, liang2024image} due to their content adaptivity and global modeling capabilities but are limited by their quadratic time complexity.
To effectively adapt to intricate image structures during pre-training, we propose a novel content-aware masked convolutional gating (MCG) mechanism to simplify attention mechanism, as depicted in Fig.~\ref{fig:network}(a). As suggested by Zhang \etal~\cite{zhang2021out}, our method initially concentrates on local features instead of global contexts to enhance efficiency and generalization. We employ masked depth-wise convolution with a $k\times k$ kernel to aggregate local contextual information. Furthermore, we simulate self-attention using gating mechanism, by modulating the value matrix through a Hadamard product with the output of the convolutional operation. The MCG mechanism can be expressed as:
\begin{equation}
\begin{aligned}
    \mathbf{A}_{\mathbf{M}} = \text{DWC}&\text{onv}_{k\times k}(\mathbf{W_A} \mathbf{X}, \mathbf{M}),  \\
\mathbf{V} &= \mathbf{W_V} \mathbf{X}, \\
\text{MCG}(\mathbf{X}) &= \sigma (\mathbf{A}_{\mathbf{M}}) \odot \mathbf{V},
\end{aligned}
\end{equation}
where $\mathbf{W_A}, \mathbf{W_V}$ are two linear projections, $\sigma$ is a non-linear activation, using swish function, $\odot$ represents the Hadamard product, and $\mathbf{M}$ is the convolutional mask that restricts information to the decoded contextual scope. 

In MCG, the \(\mathbf{A_M}\) aggregates contextual information from local regions, capturing the characteristics of local structures and details. The resulting \(\mathbf{A_M}\) is then converted into gating values through a non-linear activation function. The value matrix \(\mathbf{V}\) is a linear projection of the input, retaining the projected spatial information at each position. Content awareness is primarily achieved through the modulation operation, which regulates the flow of information within the value matrix $\mathbf{V}$ using locally adaptive gating values $\sigma(\mathbf{A_M})$. This mechanism enables the network to learn efficient adaptation to the content on training dataset, enhancing the pre-trained model's ability to respond to varying contexts.

\subsubsection{Masked Gated ConvFormer}

Building on the MCG, we introduce an autoregressive architecture called the Masked Gated ConvFormer (MGCF) inspired by the MetaFormer architecture~\cite{Yu2022Metaformer}, as illustrated in Fig.~\ref{fig:network}(b). %We are inspired by , another crucial component for the effectiveness of transformers. 
In MGCF, the input image first passes through an embedding layer, implemented via a $3\times 3$ masked convolution. This is followed by $N$ MGCF blocks, each consisting of a masked convolutional gated block, which includes MCG, and an MLP block, featuring two linear layers with GELU activation. The final output is transformed into parameters using a $1\times 1$ convolutional layer. We model with a discrete logistic mixture model, as shown in~\cite{salimans2016pixelcnn++}.

\subsubsection{Cache then Crop Inference}
To make MGCF a practical codec, we introduce a strategy to accelerate our coding process, named Cache then Crop Inference (CCI), illustrated in Fig.~\ref{fig:network} (c). CCI minimizes computations for decoded elements by caching their activations in masked convolutional layers. It then applies zero-padding convolution only to cropped windows surrounding the activated positions.

Specifically, image is first divided into patches and the patches are encoded in parallel. For each patch, pixels are initially grouped as $\bm{x} = \{\bm{x}_{\mathcal{G}_1}, \dots, \bm{x}_{\mathcal{G}_g}\}$ according to a parallel scan order, as in Fig.~\ref{fig:network} (c). This parallel scan takes inspiration from DLPR~\cite{bai2024deep} and entails $3P-2$ autoregressive steps, where $P$ is the patch size.
Encoding is conducted group by group. At step $i$, the network processes activations from previous groups $\bm{x}_{\mathcal{G}_{\leq i-1}}$ and predicts the distribution for the current group $\bm{x}_{\mathcal{G}_i}$. 
In MGCF, types A and B masked convolutions are employed. For type B convolution, current group positions are masked out. Conversely, for type A convolution, positions for the current group and all previous groups are included in the receptive field. 
For type B convolution, which is used at the top as an embedding layer, pixels from the previous group $\bm{x}_{\mathcal{G}_{i-1}}$ are fed into the network and cached. The cached data then undergoes type B masked convolution at positions corresponding to the current group, aggregating contextual information. Instead of convolving over all positions, we crop windows around the current positions and perform zero-padding convolution on these windows in parallel. In the embedding layer, activations for the current group $\bm{x}_{\mathcal{G}_i}$ are obtained and fed into the subsequent network layers.
For type A convolution, used in subsequent masked convolution gating blocks, elements at positions for the current group are considered. We store activations at current group positions into cache map and perform cropped convolution to collect contextual information for these positions. The $1\times 1$ convolution layer in the rest of the network transfers information solely across the channel dimension, and no caching is necessary.

\subsection{MDL-Principled RPFT}

\subsubsection{Low-rank Decompostion for Weights \label{subsubsec:mplora}}

We adapt the pre-trained weights using PETL to capture unique characteristics of each individual image.
Based on the hypothesis that changes in weight matrices exhibit low-rank characteristics, LoRA proposes an incremental update of pre-trained weights in linear layers using the product of two low-rank matrices. For a pre-trained weight matrix $\mathbf{W}\in \mathbb{R}^{m\times n}$, the weight update is modeled as $\Delta \mathbf{W} \in \mathbb{R}^{m\times n}$. The fine-tuned weight matrix $\mathbf{W'}$ is then given by:
\begin{equation}
    \label{eq:lora_linear}
    \mathbf{W'} = \mathbf{W} + \Delta \mathbf{W} = \mathbf{W} + \mathbf{A} \mathbf{B},
\end{equation}
where $\mathbf{A}\in\mathbb{R}^{m\times r}$ and $\mathbf{B}\in \mathbb{R}^{r\times n}$, with $r \ll \min(n, m)$. 

Additionally, aiming to fine-tuning MGCF, we extend this concept from two-dimensional linear layers to three-dimensional masked depth-wise convolutions. For a masked depth-wise convolution, the pre-trained kernel $\mathbf{W_{mc}}$ is represented in $\mathbb{R}^{m\times 1\times k\times k}$, where $k$ denotes the kernel size. Utilizing Tucker decomposition~\cite{tucker1966some}, we decompose the pre-trained weights into four components: a core identity tensor $\mathbf{I}\in \mathbb{R}^{r_1\times 1\times r_2 \times r_3}$ and three mode matrices $\mathbf{A}\in \mathbb{R}^{m\times r_1}$, $\mathbf{C}\in \mathbb{R}^{k\times r_2}$, and $\mathbf{D}\in \mathbb{R}^{k\times r_3}$. The incremental update is then defined as $\Delta \mathbf{W}_{mc}= \mathbf{I} \times_1 \mathbf{A} \times_3 \mathbf{C} \times_4 \mathbf{D}$, where $\times_n$ denotes the mode-n product, which is the multiplication of a tensor by a matrix along the n-th mode. The fine-tuned weight matrix is computed as:
\begin{equation}
    \label{eq:lora_conv}
    \mathbf{W'_{mc}}  =  \mathbf{M} \odot (\mathbf{W_{mc}} + \Delta \mathbf{W_{mc}}).
\end{equation}

Notably, the incremental weights for both linear and convolutional layers can be merged with the pre-trained weights, resulting in no additional computation time during inference. We incorporate low-rank decompostion into the masked convolutional gating blocks and MLP. Specifically, we add incremental weights to the weight matrices $\mathbf{W_A}$ and $\mathbf{W_V}$ within the convolutional gating mechanism, as well as to the first linear layer $\mathbf{W_{up}}$ in the MLP.

\subsubsection{Rate-guided Progressive Fine-Tuning \label{subsubsec:mplora}}

To further accelerate the adaptation process, we propose a Rate-guided Progressive Fine-Tuning (RPFT) approach. This method is based on the assumption that it is unnecessary to use all image patches for model fine-tuning throughout the entire process. Two key considerations support this: 1) There is inherent redundancy within patches due to content similarities among them; 2) The distributions of different patches vary in their complexity, which affects the model’s learning process differently. These considerations guide our two design aspects: first, progressively fine-tuning on an increasing number of patches, and second, determining the focus of fine-tuning—patches, which contain more informative structures or details, should be trained with more steps.

Consequently, we select the estimated bitrate as an indicator of the informational content of each patch and employ a strategy of gradually increasing the number of training patches, which are sorted in descending order by pixel rate. We begin by estimating the bitrates for all image patches, organizing them from highest to lowest, and progressively incorporate them into our training samples for fine-tuning.
Initially, we select only b\% of the patches as training samples. As training progresses, we systematically include more patches, eventually utilizing the entire set in the final d\% of the training steps. The tuning on full set ensures that our training and testing objectives are well-aligned.
We describe the process of increasing the training sample ratio $F(t)$ with the following function:
\begin{equation}
    \begin{aligned}
    t' &= \frac{t}{T \cdot (1 - d)}, \\
    s(x) &= 
    \begin{cases}
    0, & \text{if } x < 0, \\
    1, & \text{if } x > 1, \\
    x^2 (3 - 2x), & \text{if } 0 \leq x \leq 1,
    \end{cases} \\
    F(t) &= b + (1 - b) \cdot [s(t')]^e,
\end{aligned}
\end{equation}
where $t$ represents the current step, $T$ denotes the total number of training steps, and $e$ is a hyperparameter that controls the growth rate of the training sample ratio. 

\begin{figure}[t]
 \centering
 \includegraphics[width=\linewidth]{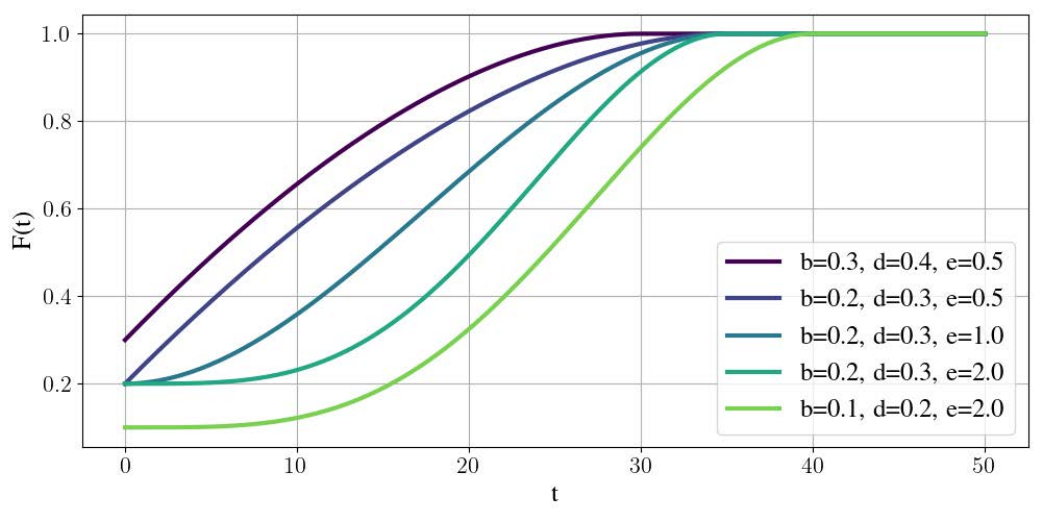}
 \caption{Illustration of incremental curves, $T=50$. The shape of the curve is controlled by parameters $b, d, e$. }
\label{fig:inc_curves}
% \vspace{-8pt}
\end{figure}

The progression of $F(t)$ is illustrated in Fig.~\ref{fig:inc_curves}.
By modifying $b$, $d$, and $e$, we can adjust the number of patches involved in the fine-tuning process. Specifically, as both $b$ and $d$ increases and $e$ decrease, the number of patches participating in training will increase. Through RPFT, we intensify our focus on learning the distribution of patches with higher informational content while effectively minimizing the number of patches involved in the training process. This strategy not only optimizes the learning process but also effectively reduces the time required for adaptation.

\begin{table*}[!t]
    \small
    \setlength{\tabcolsep}{12pt}
    \centering
    \begin{tabular}{llllllc}
    \toprule
      Codec    & Kodak & RS19 & Histo24 & DIV2K  & CLIC.p & Params. \tabularnewline
    \midrule
    PNG                                 & 4.35 \gain{+71.3\%}  & 3.90 \gain{+124.1\%}&3.79 \gain{+38.3\%} & 4.23 \gain{+72.0\%}& 3.93 \gain{+70.9\%}& $-$  \tabularnewline
        % JPEG-LS    & 3.16 \gain{+23.4\%} &   & 3.39 \gain{+27.0\%} & 2.99 \gain{+24.5\%} & 2.82 \gain{+21.5\%}  & $-$   \tabularnewline
        JPEG2000       & 3.19 \gain{+25.6\%} & 2.57 \gain{+47.7\%} & 3.36 \gain{+22.6\%} & 3.12 \gain{+30.5\%} & 2.93 \gain{+32.0\%} & $-$   \tabularnewline
        % BPG                        & 4.69 \gain{+33.1\%} &  & 3.82 \gain{+39.4\%} & 3.28 \gain{+37.2\%} &  3.08 \gain{+38.7\%} & $-$  \tabularnewline
        FLIF         & 2.90 \gain{+14.6\%} & 2.18 \gain{+25.3\%} & 3.23 \gain{+17.9\%} & 2.91 \gain{+18.3\%} & 2.72 \gain{+18.3\%} & $-$  \tabularnewline
        JPEG-XL    & 2.87 \gain{+13.4\%} & 2.02 \gain{+16.1\%} & 3.07 \gain{+12.0\%} & 2.79 \gain{+13.4\%} & 2.63 \gain{+14.3\%} & $-$  \tabularnewline
    \midrule
        L3C (CVPR'19)        & 3.26 \gain{+28.3\%} & 2.66 \gain{+53.5\%} & 3.53 \gain{+28.8\%} & 3.09 \gain{+25.6\%}  &  2.94 \gain{+27.8\%} & 5M   \tabularnewline
        RC (CVPR'20)         & 3.38 \gain{+32.8\%} & 2.19 \gain{+25.9\%} & 3.33 \gain{+21.5\%} & 3.08 \gain{+25.2\%} &  2.93 \gain{+27.4\%} & 6.9M     \tabularnewline
        iVPF (CVPR'21)           & $-$ & $-$ & $-$ & 2.68 \gain{+8.9\%}  & 2.54 \gain{+10.4\%} & 59.5M   \tabularnewline
        LC-FDNet (CVPR'22)          & 2.98 \gain{+17.3\%} & 2.15 \gain{+23.6\%} & 3.07 \gain{+12.0\%} & 2.72 \gain{+10.6\%}  & 2.63 \gain{+14.3\%} & 23.7M \tabularnewline
        DLPR (TPAMI'24)             & 2.86  \gain{+14.3\%} & 2.01 \gain{+15.5\%} & 2.96 \gain{+8.0\%} & 2.55 \gain{+3.7\%} & 2.38 \gain{+3.5\%}  & 22.2M \tabularnewline
        ArIB-BPS (CVPR'24) & 2.78 \gain{+9.8\%}& \underline{1.92} \gain{+10.3\%} &2.92 \gain{+6.6\%}&2.55 \gain{+3.7\%}&2.42 \gain{+5.2\%}& 146.6M \tabularnewline
    \midrule
        MGCF (Ours)               & \underline{2.77} \gain{+9.1\%} & 1.94 \gain{+11.5\%} & \underline{2.88} \gain{+5.1\%} & \underline{2.49} \gain{+1.2\%}   & \underline{2.33} \gain{+1.3\%} & \textbf{575K}\tabularnewline
        CALLIC (Ours)         & \textbf{2.54} & \textbf{1.74} & \textbf{2.74} & \textbf{2.46} & \textbf{2.30}  &\textbf{575K}\tabularnewline
    \bottomrule
    \end{tabular}
    \caption{Compression performance of the proposed methods and other codecs in terms of bits per sub-pixel (bpsp). We show the difference in percentage to our approach, using {\color{OliveGreen}green}. The best is highlighted in \textbf{bold}, and the second is highlighted using \underline{underline}. CALLIC introduces 25K additional trainable parameters compared to MGCF, which can be merged with the pre-trained weights during inference. } 
    \label{tb:results_ll}
    % \vspace{-10pt}
    \end{table*}

\subsubsection{MDL-Principled Optimization}

The complete lossless image compression pipeline is detailed in Fig.~\ref{fig:network} (d). Before encoding the test image into a bitstream, we first fine-tune the incremental weights for the current image as a model prompt while keeping the pre-trained MGCF parameters fixed. During encoding, these incremental parameters are quantized into discrete values before being incorporated into the bitstream. Given the relatively narrow numerical range of incremental weights, we employ a smaller quantization step size, $w < 1$. For the differentiable estimation of quantization during optimization, we utilize a mixed quantization approach, as detailed in~\cite{tsubota2023universal}. Let $\bphi$ represent the incremental weights, and $\btheta$ the pre-trained MGCF weights. 
% This process involves quantizing incremental weights using a straight-through estimator (STE)~\cite{theis2016lossy} for network inference $\hphi = \mathrm{sg}(\lfloor\bphi / w \rceil * w - \bphi) + \bphi$, where $\mathrm{sg}$ denotes stop gradient operation, while adding uniform noise to the weights for the entropy model $\tphi = \bphi + U(-\frac{w}{2}, \frac{w}{2})$.
This process quantizes incremental weights using a straight-through estimator (STE)~\cite{theis2016lossy} for network inference, where $\mathrm{sg}$ denotes stop gradient operation: $\hphi = \mathrm{sg}(\lfloor\frac{\bphi}{w} \rceil * w - \bphi) + \bphi$. Additionally, uniform noise is added to the weights for entropy estimation: $\tphi = \bphi + U(-\frac{w}{2}, \frac{w}{2})$.
We leverage the MDL principle to jointly optimize the bitrates for both the incremental weights and the image pixels:
\begin{equation}
    \mathcal{L} = - \log p_s(\tphi) + \sum_i - \log q(x_{\mathcal{G}_i} \mid x_{\mathcal{G}_{<i}} ; \btheta, \hphi),
\end{equation}
where $p_s(\tphi)$ denotes the probability distribution of the incremental weights, modeled using a static logistic distribution with zero mean and a constant scale $ s \in \mathbb{R} $.

After optimization, we explicitly compute and merge the weights according to Eq.~\ref{eq:lora_linear} and Eq.~\ref{eq:lora_conv}, and then proceed with network inference as usual, ensuring no additional time is required during this stage. Finally, the image is encoded into a bitstream using the final model. The total bitrate for compressing the image includes the bitrates for the quantized incremental weights and the image pixels. This bitstream is then transmitted to the decoder, where the incremental weights are first decoded and used for image decoding.

\begin{table}[t!]
    \centering
    \small 
    % \fontsize{9pt}{11pt}\selectfont
    % \setlength{\tabcolsep}{1mm} 
    \begin{tabular}{lccccc}
    \toprule
     Codec & Bpsp & Params.&  Enc. Time & Dec. Time \\
        %  \midrule
        %   BPG & 3.38 & - & 2.4s & 0.1s \\
         \midrule
          LC-FDNet & 2.98 & 23.7M & 1.6s & 1.6s \\
          DLPR & 2.86 & 22.2M & 1.5s & 2.0s  \\
          ArIB-BPS & 2.78 & 146.6M & 7.1s & 7.0s \\
          \midrule
          MGCF (Ours) & 2.77 & 575K & 1.4s & 1.6s \\
          \textit{-w/o CCI} & $-$ & $-$ & \textit{15.2s} & \textit{15.4s} \\
          CALLIC (Ours) & 2.54 & 575K & 9.7s & 1.7s \\
          \textit{-w/o RPFT} & \textit{2.54} & $-$ & \textit{12.9s} & $-$ \\
        %   MGCF (fp16) & 2.77 & - & 1.3s & 1.5s \\
        %   CALLIC & 2.56 & 790K & 93.7s  & 7.7s   \\
         \bottomrule
    \end{tabular}
    \caption{Coding speed analysis on Kodak. Bpsp denotes bits per sub-pixel. Our pre-trained model delivers fast coding speeds, leveraging the proposed CCI mechanism. Fine-tuning improves compression but increases encoding time, which is effectively reduced with the RPFT method.}
    \label{tb:results_time}
    % \vspace{-5pt}
\end{table}

\begin{figure}[t]
 \centering
 \includegraphics[width=\linewidth]{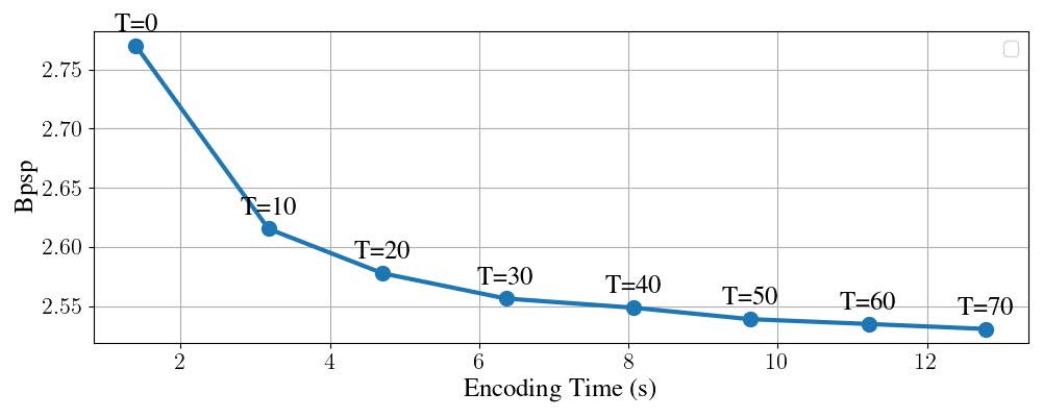}
 \caption{Steps and performance trade-off for CALLIC. Faster encoding can be achieved with minimal performance trade-offs.}
\label{fig:exp_steps}
% \vspace{-10pt}
\end{figure}

\section{Experiments}

\subsection{Experimental Settings}
% \subsubsection{Network Settings}
% In the implementation details of the MGCF network structure, we opt for a  depth $N$ of $3$, a hidden dimension of $128$, and kernel size is $7$ for depth-wise convolution, MLP ratio of $4$. 
% The rank for MPLoRA is set to $r=8$. 
% We employ \textit{torchac}~\cite{mentzer2019practical}, an arithmetic coding library, for entropy coding.
\subsubsection{Pre-trained Settings}
For pretraining, we make a collection of $3450$ images from publicly available high-resolution datasets: DIV2K~\cite{Agustsson_2017_CVPR_Workshops}, with $800$ training images, and Flickr2K~\cite{Lim_2017_CVPR_Workshops}, with $2650$ training images. We crop these images into non-overlapping patches of size $64\times 64$, resulting in a training set of $612,806$ images. We train MGCF for 2M steps using the Adam optimizer with minibatches of size $32$ and learning rate $5e-4$. 
% We use cosine learning rate scheduler, and the initial learning rate is set to $1e-3$. 
\subsubsection{Adaptation Settings}
The maximum number of optimization steps is set to $T=50$, with a learning rate $1e-2$. A relatively small scale for the prior weight distribution is chosen, set as $s=0.05$. Quantization width is set to $w=0.05$. For RPFT, we choose $b=0.2,d=0.1,e=1$ as default.

\subsubsection{Evaluation Settings}
We selected five high-resolution datasets: Kodak~\cite{kodak}, WHU-RS19 validation~\cite{Xia2010WHURS19}, Histo24~\cite{bai2024deep}, DIV2K~\cite{Agustsson_2017_CVPR_Workshops} and CLIC2020 professional validation (CLIC.p)~\cite{clic} as our evaluation datasets. The Kodak, DIV2K, CLIC.p, and CLIC.m are natural images datasets. Histo24 is a dataset that includes 24 $768\times 512$ histological images proposed by~\cite{bai2024deep}. We center-cropped 190 satellite images to $576\times 576$ from the WHU-RS19 validation set, which were exported from Google Earth, to form our validation dataset, RS19. 
Our method is compared with traditional lossless codecs, including JPEG2000 \cite{skodras2001jpeg}, FLIF \cite{sneyers2016flif}, and JPEG-XL \cite{alakuijala2019jpeg}, and \textit{open-sourced} learned methods like L3C~\cite{mentzer2019practical}, RC~\cite{mentzer2020learning}, iVPF~\cite{zhang2021ivpf}, LC-FDNet~\cite{rhee2022lc}, DLPR~\cite{bai2024deep} and ArIB-BPS~\cite{Zhang2024ArIBBPS}.
% \vspace{-5pt}

\subsection{Experimental Results}
\subsubsection{Model Performance}
The compression performance results are summarized in Tab.~\ref{tb:results_ll}. Our pre-trained model, MGCF, demonstrates superior performance compared to previous methods. Specifically, MGCF outperforms all other methods in four out of five datasets and performs slightly worse than ArIB-BPS on the RS19 dataset.
By integrating RPFT into MGCF during encoding to create CALLIC, we achieve further improvements. CALLIC consistently surpasses all other methods across all datasets. 
For datasets with gaps between the training set, such as Kodak, RS19, and Histo24, CALLIC achieves significant improvements of 9.1\%, 11.5\%, and 5.1\% over MGCF, respectively. For other datasets, CALLIC also shows better, albeit less significant, improvements. This is attributed to the effectiveness of MGCF, which has learned similar distributions on the training dataset, making further gains challenging.
We highlight that MGCF and CALLIC are lightweight models, with MGCF having only 575K parameters and CALLIC adding 25K \textit{mergeable}  weights. This is significantly smaller than other methods. Compared to ArIB-BPS, MGCF achieves better or comparable results with only \textbf{4\textperthousand} of the parameters.
%  The lightweight nature of these models ensures the feasibility of further adaptation.
\textit{It is worth noting that the lightweight characteristic is what makes further adaptation possible and efficient.}

\begin{table}[t]
    \centering
    \small
    \begin{tabular}{cccccc}
    \toprule
     Depth & Dim. & Kernel  & Bpsp  &Enc./Dec. Time \\
         \midrule
        %  \multicolumn{5}{l}{\textit{Depth}} \\
          1 & 128 & 7& 2.84&  1.0/1.1s  \\
          \textbf{3} & \textbf{128} & \textbf{7}& \textbf{2.77} &  \textbf{1.4/1.6s} \\
          5 & 128 & 7 & 2.73 &  2.0/2.1s \\
         \midrule
        %  \multicolumn{5}{l}{\textit{Dimension}} \\
          3 & 64 & 7 & 2.82  & 1.4/1.5s  \\
          % \textbf{3} & \textbf{128} & \textbf{7}& \textbf{2.77} &  \textbf{1.4/1.6s} \\
          3 & 192 & 7 & 2.74 & 1.8/1.8s \\
          3 & 256 & 7 & 2.72 & 2.1/2.1s \\
         \midrule
        %  \multicolumn{5}{l}{\textit{Kernei Size}} \\
        %   3 & 128 & 3 & 2.86  & 1.4/1.5s  \\
          3 & 128 & 5 & 2.82 & 1.4/1.5s \\
          % \textbf{3} & \textbf{128} & \textbf{7}& \textbf{2.77} &  \textbf{1.4/1.6s} \\
          3 & 128 & 9 & 2.77 & 1.8/1.9s \\
          3 & 128 & 11 & 2.76 & 2.2/2.3s \\
         \bottomrule
    \end{tabular}
    \caption{Analysis about network architecture. We perform an ablation study on network depth, model channels, and the kernel size of masked convolutions. }
    \label{tab:ab_network}
    % \vspace{-8pt}
\end{table}

\subsubsection{Runtime}

We present the encoding speeds for recent learned methods on the Kodak dataset in Tab.~\ref{tb:results_time}. MGCF achieves competitive encoding and decoding times of 1.4 and 1.6 seconds, respectively, which are faster or comparable to other advanced methods. When the CCI is removed, encoding and decoding times increase to 15.2 and 15.4 seconds respectively.
Content adaptive fune-tuning enhances performance while increasing the encoding time. The introduction of RPFT reduces the encoding time from 12.9 to 9.7 seconds while maintaining same performance, demonstrating its effectiveness.
A trade-off between speed and performance can be precisely adjusted by modifying the training steps, as illustrated in Fig~\ref{fig:exp_steps}. For instance, with $T=10$, the encoding time is reduced to 3.2 seconds, with a bpsp of 2.62.

\begin{table}[t]
    \centering
    
    \small 

    \begin{tabular}{lccccc}
    \toprule
     Method & Bpsp & Params. & Enc. Time & Dec. Time \\
         \midrule
          MGCF & 2.77 & 575K & 1.4s & 1.6s \\
          NAT & 2.78 & 767K & 7.6s & 7.8s \\
         \bottomrule
    \end{tabular}
    \caption{Comparison with neighborhood attention mechanism on Kodak. Masked convolutional gating mechanism performs better than neighborhood attention mechanism. }
    \label{tb:compare_with_nat}
    % \vspace{-10pt}
\end{table}

\subsection{Ablation Studies \label{subsec:ablation_studies}}
\subsubsection{Network Architecture of MGCF}
The Tab.~\ref{tab:ab_network} presents an analysis of the impact of various hyperparameters on the performance of MGCF. The study explores changes in model complexity by varying the depth, dimension, and kernel size. The results demonstrate that increasing model complexity generally enhances performance. Notably, while increasing the kernel size beyond 7 shows diminishing returns, expanding the dimension and depth consistently leads to performance gains. The final model, with a depth of 3, a dimension of 128, and a kernel size of 7, was chosen to balance the trade-off between improved performance and competitive encoding/decoding complexity.

\subsubsection{Convolutional Gating vs. Neighborhood Attention}
To compare the effectiveness of original self-attention and convolutional gating mechanisms, we adopt Neighborhood Attention Transformer~\cite{hassani2023neighborhood}, which uses local attention on feature maps within a sliding window. In our experiments, we replaced masked convolutional gating with masked neighborhood attention, called ``NAT". The results, shown in Tab.~\ref{tb:compare_with_nat}, indicate that convolutional gating not only performs better but also works faster, mainly because it uses only convolution operations.

\begin{figure}[t]
 \centering
 \includegraphics[width=\linewidth]{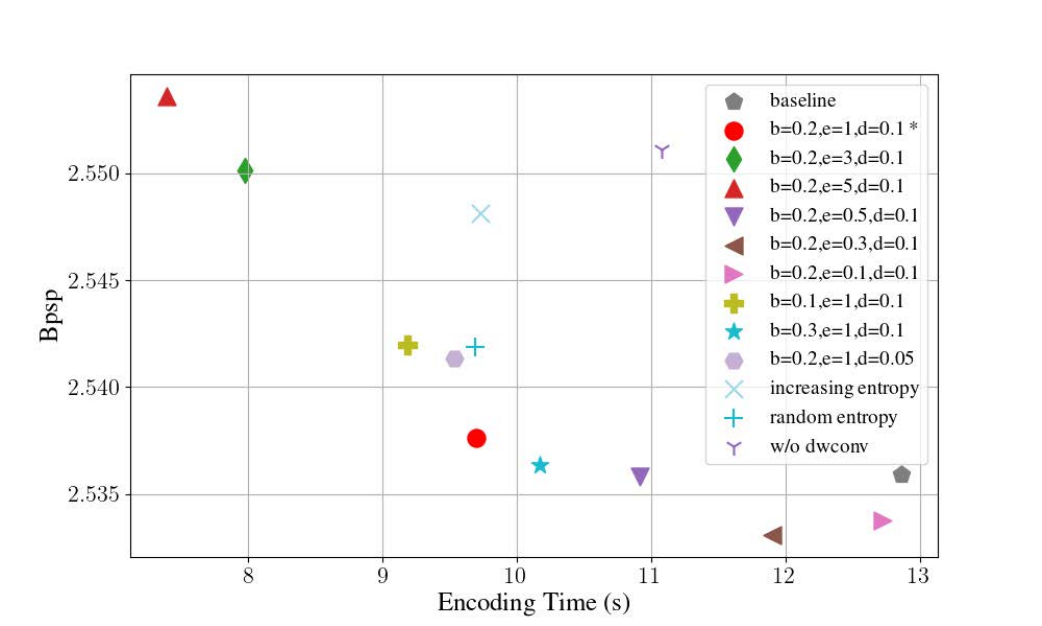}
 \caption{Different configurations for RPFT on Kodak. We ablate different increasing curves and sorting orders in this experiment.}
\label{fig:ab_RPFT}
% \vspace{-10pt}
\end{figure}

\subsubsection{Different Configurations for RPFT}

The experimental data presented in Fig.~\ref{fig:ab_RPFT} explores the effects of different configurations for RPFT. The baseline method, without RPFT, results in an encoding time of 12.9 seconds and a bpsp of 2.535. When employing RPFT with default configuration (b=0.2, e=1, d=0.1), the encoding time significantly decreases to 9.7 seconds while maintaining a comparable bpsp of 2.537. We found that the default configuration demonstrates robust performance across different image types, making it a reliable choice for various datasets.

Increasing the participation of image patches, characterized by higher b and d values and lower e values, generally leads to improved performance, albeit at the cost of longer encoding times. 
% Conversely, slower rates result in shorter encoding times but reduced performance.
Notably, the configurations (b=0.2, e=0.3, d=0.1) and (b=0.2, e=0.1, d=0.1) outperform the baseline in both encoding time and compression performance, underscoring the effectiveness of RPFT. 

The results in Fig.~\ref{fig:ab_RPFT} also indicate that focusing on higher rate patches (default) achieves better performance compared to starting from lower rate patches (``increasing entropy") or employing random sampling (``random entropy") strategies. 

The absence of adaptation for depth-wise convolutions, as seen in the ``w/o dwconv" condition, results in a higher bpsp of 2.551. This highlights the crucial role of depth-wise convolution adaptation in enhancing overall performance.

\section{Conclusion}
% In this work, we propose CALLIC, by exploring the connection between the MDL principle and PETL. We first introduce an efficient pre-trained autoregressive model MGCF with an acceleration strategy CCI to make it a practical codec. Then, we decompose layers including depth-wise convolution using low-rank matrices and propose RPFT to efficiently fine-tune MGCF on testing image. Extensive experiments demonstrate that CALLIC sets a new SOTA.
In this work, we propose CALLIC by leveraging the connection between the MDL principle and PETL. We introduce MGCF, an efficient pre-trained autoregressive model, combined with the CCI acceleration strategy to enhance its practicality as a codec. For efficient fine-tuning on test images, we decompose layers, including depth-wise convolutions, using low-rank matrices and introduce RPFT. This approach progressively fine-tunes additional parameters using gradually increasing patches prioritized by estimated entropy. Extensive experiments demonstrate that CALLIC establishes a new SOTA in learned lossless image compression.

\section{Acknowledgments}
% This work was supported in part by National Key Research and Development Program of China under Grant 2022YFF1202104, in part by National Natural Science Foundation of China under Grants 62301188, 92270116 and U23B2009, in part by China Postdoctoral Science Foundation under Grant 2022M710958, and in part by Heilongjiang Postdoctoral Science Foundation under Grant LBH-Z22156.
This work was supported in part by National Key Research and Development Program of China under Grant 2022YFF1202104, in part by the National Key R\&D Program of China under Grant 2023YFC2509100, in part by National Natural Science Foundation of China under Grants 62301188, 92270116 and U23B2009, in part by the Strategic Research, and Consulting Project by the Chinese Academy of Engineering under Grant 2023.XY-39, in part by China Postdoctoral Science Foundation under Grant 2022M710958, and in part by Heilongjiang Postdoctoral Science Foundation under Grant LBH-Z22156.

\bibliography{aaai25}

\begin{thebibliography}{49}
\providecommand{\natexlab}[1]{#1}

\bibitem[{Agustsson and Timofte(2017)}]{Agustsson_2017_CVPR_Workshops}
Agustsson, E.; and Timofte, R. 2017.
\newblock NTIRE 2017 Challenge on Single Image Super-Resolution: Dataset and Study.
\newblock In \emph{The IEEE Conference on Computer Vision and Pattern Recognition (CVPR) Workshops}.

\bibitem[{Alakuijala et~al.(2019)Alakuijala, Van~Asseldonk, Boukortt, Bruse, Comșa, Firsching, Fischbacher, Kliuchnikov, Gomez, Obryk et~al.}]{alakuijala2019jpeg}
Alakuijala, J.; Van~Asseldonk, R.; Boukortt, S.; Bruse, M.; Comșa, I.-M.; Firsching, M.; Fischbacher, T.; Kliuchnikov, E.; Gomez, S.; Obryk, R.; et~al. 2019.
\newblock JPEG XL next-generation image compression architecture and coding tools.
\newblock In \emph{Applications of digital image processing XLII}, volume 11137, 112--124. SPIE.

\bibitem[{Bai et~al.(2024)Bai, Liu, Wang, Ji, Wu, and Gao}]{bai2024deep}
Bai, Y.; Liu, X.; Wang, K.; Ji, X.; Wu, X.; and Gao, W. 2024.
\newblock Deep lossy plus residual coding for lossless and near-lossless image compression.
\newblock \emph{IEEE Transactions on Pattern Analysis and Machine Intelligence}.

\bibitem[{Bai et~al.(2021)Bai, Liu, Zuo, Wang, and Ji}]{bai2021learning}
Bai, Y.; Liu, X.; Zuo, W.; Wang, Y.; and Ji, X. 2021.
\newblock Learning scalable l$_\infty$-constrained near-lossless image compression via joint lossy image and residual compression.
\newblock In \emph{Proceedings of the IEEE/CVF Conference on Computer Vision and Pattern Recognition}, 11946--11955.

\bibitem[{Bai et~al.(2022)Bai, Yang, Liu, Jiang, Wang, Ji, and Gao}]{bai2022towards}
Bai, Y.; Yang, X.; Liu, X.; Jiang, J.; Wang, Y.; Ji, X.; and Gao, W. 2022.
\newblock Towards end-to-end image compression and analysis with transformers.
\newblock In \emph{Proceedings of the AAAI conference on artificial intelligence}, volume~36, 104--112.

\bibitem[{Ball{\'e}, Laparra, and Simoncelli(2017)}]{balle2017end}
Ball{\'e}, J.; Laparra, V.; and Simoncelli, E.~P. 2017.
\newblock End-to-end optimized image compression.
\newblock In \emph{5th International Conference on Learning Representations, ICLR 2017}.

\bibitem[{Barron, Rissanen, and Yu(1998)}]{barron1998minimum}
Barron, A.; Rissanen, J.; and Yu, B. 1998.
\newblock The minimum description length principle in coding and modeling.
\newblock \emph{IEEE transactions on information theory}, 44(6): 2743--2760.

\bibitem[{Cheng et~al.(2020)Cheng, Sun, Takeuchi, and Katto}]{cheng2020learned}
Cheng, Z.; Sun, H.; Takeuchi, M.; and Katto, J. 2020.
\newblock Learned image compression with discretized gaussian mixture likelihoods and attention modules.
\newblock In \emph{Proceedings of the IEEE/CVF conference on computer vision and pattern recognition}, 7939--7948.

\bibitem[{Dosovitskiy(2020)}]{dosovitskiy2020image}
Dosovitskiy, A. 2020.
\newblock An image is worth 16x16 words: Transformers for image recognition at scale.
\newblock \emph{arXiv preprint arXiv:2010.11929}.

\bibitem[{Gr{\"u}nwald(2005)}]{grunwald2005minimum}
Gr{\"u}nwald, P. 2005.
\newblock Minimum description length tutorial.

\bibitem[{Hassani et~al.(2023)Hassani, Walton, Li, Li, and Shi}]{hassani2023neighborhood}
Hassani, A.; Walton, S.; Li, J.; Li, S.; and Shi, H. 2023.
\newblock Neighborhood attention transformer.
\newblock In \emph{Proceedings of the IEEE/CVF Conference on Computer Vision and Pattern Recognition}, 6185--6194.

\bibitem[{He et~al.(2022)He, Yang, Peng, Ma, Qin, and Wang}]{he2022elic}
He, D.; Yang, Z.; Peng, W.; Ma, R.; Qin, H.; and Wang, Y. 2022.
\newblock Elic: Efficient learned image compression with unevenly grouped space-channel contextual adaptive coding.
\newblock In \emph{Proceedings of the IEEE/CVF Conference on Computer Vision and Pattern Recognition}, 5718--5727.

\bibitem[{Ho, Lohn, and Abbeel(2019)}]{ho2019compression}
Ho, J.; Lohn, E.; and Abbeel, P. 2019.
\newblock Compression with flows via local bits-back coding.
\newblock \emph{Advances in Neural Information Processing Systems}, 32.

\bibitem[{Hoogeboom et~al.(2019)Hoogeboom, Peters, Van Den~Berg, and Welling}]{hoogeboom2019integer}
Hoogeboom, E.; Peters, J.; Van Den~Berg, R.; and Welling, M. 2019.
\newblock Integer discrete flows and lossless compression.
\newblock \emph{Advances in Neural Information Processing Systems}, 32.

\bibitem[{Houlsby et~al.(2019)Houlsby, Giurgiu, Jastrzebski, Morrone, De~Laroussilhe, Gesmundo, Attariyan, and Gelly}]{houlsby2019parameter}
Houlsby, N.; Giurgiu, A.; Jastrzebski, S.; Morrone, B.; De~Laroussilhe, Q.; Gesmundo, A.; Attariyan, M.; and Gelly, S. 2019.
\newblock Parameter-efficient transfer learning for NLP.
\newblock In \emph{International conference on machine learning}, 2790--2799. PMLR.

\bibitem[{Hu et~al.(2021)Hu, Wallis, Allen-Zhu, Li, Wang, Wang, Chen et~al.}]{hu2021lora}
Hu, E.~J.; Wallis, P.; Allen-Zhu, Z.; Li, Y.; Wang, S.; Wang, L.; Chen, W.; et~al. 2021.
\newblock LoRA: Low-Rank Adaptation of Large Language Models.
\newblock In \emph{International Conference on Learning Representations}.

\bibitem[{Kingma, Abbeel, and Ho(2019)}]{kingma2019bit}
Kingma, F.; Abbeel, P.; and Ho, J. 2019.
\newblock Bit-swap: Recursive bits-back coding for lossless compression with hierarchical latent variables.
\newblock In \emph{International Conference on Machine Learning}, 3408--3417. PMLR.

\bibitem[{Kodak(1993)}]{kodak}
Kodak, E. 1993.
\newblock Kodak lossless true color image suite (PhotoCD PCD0992).
\newblock \url{http://r0k.us/graphics/kodak}.

\bibitem[{Lester, Al-Rfou, and Constant(2021)}]{lester2021power}
Lester, B.; Al-Rfou, R.; and Constant, N. 2021.
\newblock The Power of Scale for Parameter-Efficient Prompt Tuning.
\newblock In \emph{Proceedings of the 2021 Conference on Empirical Methods in Natural Language Processing}, 3045--3059.

\bibitem[{Li et~al.(2024{\natexlab{a}})Li, Bai, Wang, Jiang, and Liu}]{li2024semantic}
Li, D.; Bai, Y.; Wang, K.; Jiang, J.; and Liu, X. 2024{\natexlab{a}}.
\newblock Semantic Ensemble Loss and Latent Refinement for High-Fidelity Neural Image Compression.
\newblock \emph{arXiv preprint arXiv:2401.14007}.

\bibitem[{Li et~al.(2024{\natexlab{b}})Li, Bai, Wang, Jiang, Liu, and Gao}]{li2024groupedmixer}
Li, D.; Bai, Y.; Wang, K.; Jiang, J.; Liu, X.; and Gao, W. 2024{\natexlab{b}}.
\newblock GroupedMixer: An Entropy Model with Group-wise Token-Mixers for Learned Image Compression.
\newblock \emph{IEEE Transactions on Circuits and Systems for Video Technology}.

\bibitem[{Li and Liang(2021)}]{li2021prefix}
Li, X.~L.; and Liang, P. 2021.
\newblock Prefix-Tuning: Optimizing Continuous Prompts for Generation.
\newblock In \emph{Proceedings of the 59th Annual Meeting of the Association for Computational Linguistics and the 11th International Joint Conference on Natural Language Processing (Volume 1: Long Papers)}, 4582--4597.

\bibitem[{Liang et~al.(2024)Liang, Jiang, Liu, and Ma}]{liang2024image}
Liang, P.; Jiang, J.; Liu, X.; and Ma, J. 2024.
\newblock Image deblurring by exploring in-depth properties of transformer.
\newblock \emph{IEEE Transactions on Neural Networks and Learning Systems}.

\bibitem[{Lim et~al.(2017)Lim, Son, Kim, Nah, and Lee}]{Lim_2017_CVPR_Workshops}
Lim, B.; Son, S.; Kim, H.; Nah, S.; and Lee, K.~M. 2017.
\newblock Enhanced Deep Residual Networks for Single Image Super-Resolution.
\newblock In \emph{The IEEE Conference on Computer Vision and Pattern Recognition (CVPR) Workshops}.

\bibitem[{Mentzer et~al.(2019)Mentzer, Agustsson, Tschannen, Timofte, and Gool}]{mentzer2019practical}
Mentzer, F.; Agustsson, E.; Tschannen, M.; Timofte, R.; and Gool, L.~V. 2019.
\newblock Practical full resolution learned lossless image compression.
\newblock In \emph{Proceedings of the IEEE/CVF conference on computer vision and pattern recognition}, 10629--10638.

\bibitem[{Mentzer, Gool, and Tschannen(2020)}]{mentzer2020learning}
Mentzer, F.; Gool, L.~V.; and Tschannen, M. 2020.
\newblock Learning better lossless compression using lossy compression.
\newblock In \emph{Proceedings of the IEEE/CVF Conference on Computer Vision and Pattern Recognition}, 6638--6647.

\bibitem[{Rhee et~al.(2022)Rhee, Jang, Kim, and Cho}]{rhee2022lc}
Rhee, H.; Jang, Y.~I.; Kim, S.; and Cho, N.~I. 2022.
\newblock LC-FDNet: Learned lossless image compression with frequency decomposition network.
\newblock In \emph{Proceedings of the IEEE/CVF conference on computer vision and pattern recognition}, 6033--6042.

\bibitem[{Ryder et~al.(2022)Ryder, Zhang, Kang, and Zhang}]{ryder2022split}
Ryder, T.; Zhang, C.; Kang, N.; and Zhang, S. 2022.
\newblock Split hierarchical variational compression.
\newblock In \emph{Proceedings of the IEEE/CVF Conference on Computer Vision and Pattern Recognition}, 386--395.

\bibitem[{Salimans et~al.(2016)Salimans, Karpathy, Chen, and Kingma}]{salimans2016pixelcnn++}
Salimans, T.; Karpathy, A.; Chen, X.; and Kingma, D.~P. 2016.
\newblock PixelCNN++: Improving the PixelCNN with Discretized Logistic Mixture Likelihood and Other Modifications.
\newblock In \emph{International Conference on Learning Representations}.

\bibitem[{Shen, Yue, and Yang(2023)}]{shen2023dec}
Shen, S.; Yue, H.; and Yang, J. 2023.
\newblock Dec-adapter: Exploring efficient decoder-side adapter for bridging screen content and natural image compression.
\newblock In \emph{Proceedings of the IEEE/CVF International Conference on Computer Vision}, 12887--12896.

\bibitem[{Skodras, Christopoulos, and Ebrahimi(2001)}]{skodras2001jpeg}
Skodras, A.; Christopoulos, C.; and Ebrahimi, T. 2001.
\newblock The JPEG 2000 still image compression standard.
\newblock \emph{IEEE Signal processing magazine}, 18(5): 36--58.

\bibitem[{Sneyers and Wuille(2016)}]{sneyers2016flif}
Sneyers, J.; and Wuille, P. 2016.
\newblock FLIF: Free lossless image format based on MANIAC compression.
\newblock In \emph{2016 IEEE international conference on image processing (ICIP)}, 66--70. IEEE.

\bibitem[{Theis et~al.(2016)Theis, Shi, Cunningham, and Husz{\'a}r}]{theis2016lossy}
Theis, L.; Shi, W.; Cunningham, A.; and Husz{\'a}r, F. 2016.
\newblock Lossy image compression with compressive autoencoders.
\newblock In \emph{International Conference on Learning Representations}.

\bibitem[{Toderici et~al.(2020)Toderici, Timofte, Ballé, Agustsson, Johnston, and Mentzer}]{clic}
Toderici, G.; Timofte, R.; Ballé, J.; Agustsson, E.; Johnston, N.; and Mentzer, F. 2020.
\newblock Workshop and Challenge on Learned Image Compression (CLIC).
\newblock \url{http://www.compression.cc}.

\bibitem[{Townsend, Bird, and Barber(2019)}]{townsend2019practical}
Townsend, J.; Bird, T.; and Barber, D. 2019.
\newblock Practical lossless compression with latent variables using bits back coding.
\newblock In \emph{7th International Conference on Learning Representations, ICLR 2019}, volume~7. International Conference on Learning Representations (ICLR).

\bibitem[{Townsend et~al.(2019)Townsend, Bird, Kunze, and Barber}]{townsend2019hilloc}
Townsend, J.; Bird, T.; Kunze, J.; and Barber, D. 2019.
\newblock HiLLoC: lossless image compression with hierarchical latent variable models.
\newblock In \emph{International Conference on Learning Representations}.

\bibitem[{Tsubota, Akutsu, and Aizawa(2023)}]{tsubota2023universal}
Tsubota, K.; Akutsu, H.; and Aizawa, K. 2023.
\newblock Universal deep image compression via content-adaptive optimization with adapters.
\newblock In \emph{Proceedings of the IEEE/CVF Winter Conference on Applications of Computer Vision}, 2529--2538.

\bibitem[{Tucker(1966)}]{tucker1966some}
Tucker, L.~R. 1966.
\newblock Some mathematical notes on three-mode factor analysis.
\newblock \emph{Psychometrika}, 31(3): 279--311.

\bibitem[{Van Den~Oord, Kalchbrenner, and Kavukcuoglu(2016)}]{van2016pixel}
Van Den~Oord, A.; Kalchbrenner, N.; and Kavukcuoglu, K. 2016.
\newblock Pixel recurrent neural networks.
\newblock In \emph{International conference on machine learning}, 1747--1756. PMLR.

\bibitem[{Vaswani(2017)}]{vaswani2017attention}
Vaswani, A. 2017.
\newblock Attention is all you need.
\newblock \emph{Advances in Neural Information Processing Systems}.

\bibitem[{Wang et~al.(2024)Wang, Bai, Li, Zhai, Jiang, and Liu}]{wang2024learning}
Wang, K.; Bai, Y.; Li, D.; Zhai, D.; Jiang, J.; and Liu, X. 2024.
\newblock Learning Lossless Compression for High Bit-Depth Volumetric Medical Image.
\newblock \emph{IEEE Transactions on Image Processing}, 1--1.

\bibitem[{Wang et~al.(2023)Wang, Bai, Zhai, Li, Jiang, and Liu}]{wang2023learning}
Wang, K.; Bai, Y.; Zhai, D.; Li, D.; Jiang, J.; and Liu, X. 2023.
\newblock Learning lossless compression for high bit-depth medical imaging.
\newblock In \emph{2023 IEEE International conference on multimedia and expo (ICME)}, 2549--2554. IEEE.

\bibitem[{Weinberger, Seroussi, and Sapiro(2000)}]{weinberger2000loco}
Weinberger, M.~J.; Seroussi, G.; and Sapiro, G. 2000.
\newblock The LOCO-I lossless image compression algorithm: Principles and standardization into JPEG-LS.
\newblock \emph{IEEE Transactions on Image processing}, 9(8): 1309--1324.

\bibitem[{Xia et~al.(2010)Xia, Yang, Delon, Gousseau, Sun, and MaÎtre}]{Xia2010WHURS19}
Xia, G.-S.; Yang, W.; Delon, J.; Gousseau, Y.; Sun, H.; and MaÎtre, H. 2010.
\newblock Structural high-resolution satellite image indexing.
\newblock Vienna, Austria.

\bibitem[{Yu et~al.(2022)Yu, Luo, Zhou, Si, Zhou, Wang, Feng, and Yan}]{Yu2022Metaformer}
Yu, W.; Luo, M.; Zhou, P.; Si, C.; Zhou, Y.; Wang, X.; Feng, J.; and Yan, S. 2022.
\newblock MetaFormer Is Actually What You Need for Vision.
\newblock In \emph{Proceedings of the IEEE/CVF Conference on Computer Vision and Pattern Recognition (CVPR)}, 10819--10829.

\bibitem[{Zhang, Zhang, and McDonagh(2021)}]{zhang2021out}
Zhang, M.; Zhang, A.; and McDonagh, S. 2021.
\newblock On the out-of-distribution generalization of probabilistic image modelling.
\newblock \emph{Advances in Neural Information Processing Systems}, 34: 3811--3823.

\bibitem[{Zhang et~al.(2021{\natexlab{a}})Zhang, Kang, Ryder, and Li}]{zhang2021iflow}
Zhang, S.; Kang, N.; Ryder, T.; and Li, Z. 2021{\natexlab{a}}.
\newblock iflow: Numerically invertible flows for efficient lossless compression via a uniform coder.
\newblock \emph{Advances in Neural Information Processing Systems}, 34: 5822--5833.

\bibitem[{Zhang et~al.(2021{\natexlab{b}})Zhang, Zhang, Kang, and Li}]{zhang2021ivpf}
Zhang, S.; Zhang, C.; Kang, N.; and Li, Z. 2021{\natexlab{b}}.
\newblock ivpf: Numerical invertible volume preserving flow for efficient lossless compression.
\newblock In \emph{Proceedings of the IEEE/CVF Conference on Computer Vision and Pattern Recognition}, 620--629.

\bibitem[{Zhang et~al.(2024)Zhang, Wang, Chen, and Liu}]{Zhang2024ArIBBPS}
Zhang, Z.; Wang, H.; Chen, Z.; and Liu, S. 2024.
\newblock Learned Lossless Image Compression based on Bit Plane Slicing.
\newblock In \emph{Proceedings of the IEEE/CVF Conference on Computer Vision and Pattern Recognition (CVPR)}, 27579--27588.

\end{thebibliography}

\end{document}